\documentclass[runningheads]{llncs}
\usepackage{graphicx}
\usepackage{amsmath}
\usepackage{amssymb}
\usepackage{multirow}
\usepackage{graphicx}

\begin{document}
%
\title{Opinion Transmission Network for Jointly Improving Aspect-oriented Opinion Words Extraction and Sentiment Classification}
\titlerunning{OTN for Jointly Improving AOWE and ALSC}
%
\newcommand*\samethanks[1][\value{footnote}]{\footnotemark[#1]}
\author{Chengcan Ying\thanks{Authors contributed equally.} \and
Zhen Wu\samethanks \and
Xinyu Dai\thanks{Corresponding author.}\orcidID{0000-0002-4139-7337} \and
\\Shujian Huang \and
Jiajun Chen}
\authorrunning{C. Ying et al.}
\institute{National Key Laboratory for Novel Software Technology, Nanjing University, Nanjing, 210023, China\\
\email{\{yingcc,wuz\}@smail.nju.edu.cn},
\email{\{daixinyu,huangsj,chenjj\}@nju.edu.cn}
}
\maketitle              
\begin{abstract}
Aspect-level sentiment classification (ALSC) and aspect oriented opinion words extraction (AOWE) are two highly relevant aspect-based sentiment analysis (ABSA) subtasks. They respectively aim to detect the sentiment polarity and extract the corresponding opinion words toward a given aspect in a sentence. Previous works separate them and focus on one of them by training neural models on small-scale labeled data, while neglecting the connections between them. In this paper, we propose a novel joint model, Opinion Transmission Network (OTN), to exploit the potential bridge between ALSC and AOWE to achieve the goal of facilitating them simultaneously. Specifically, we design two tailor-made opinion transmission mechanisms to control opinion clues flow bidirectionally, respectively from ALSC to AOWE and AOWE to ALSC. Experiment results on two benchmark datasets show that our joint model outperforms strong baselines on the two tasks. Further analysis also validates the effectiveness of opinion transmission mechanisms.

\keywords{Aspect-level sentiment classification  \and Aspect-oriented opinion words extraction \and Opinion transmission network.}
\end{abstract}
\section{Introduction}
Aspect-based sentiment analysis (ABSA) is a fine-grained sentiment analysis task~\cite{pontiki-etal-2014-semeval}, which analyzes the sentiment or opinions toward a given aspect in a sentence. The task consists of a set of subtasks, including aspect category detection, aspect term extraction, aspect-level sentiment classification (ALSC), and aspect-oriented opinion words extraction (AOWE), etc. Most existing researches perform a certain subtask of ABSA through training machine learning algorithms on labeled data~\cite{wang2016attention,chen2017recurrent,xue2018aspect}. However, the public corpora of ABSA are all small-scale due to the expensive and labor-intensive manual annotation. Scarce training data limits the performance of data-driven approaches for ABSA. Therefore, an interesting and valuable research question is how to mine and exploit internal connections between ABSA subtasks to achieve the goal of facilitating them simultaneously. In this work, we focus on two subtasks ALSC and AOWE because they are highly mutually indicative. We first introduce them briefly before presenting our motivations.

\begin{figure}[!htbp]
	\centering
	\includegraphics[width=0.8\textwidth]{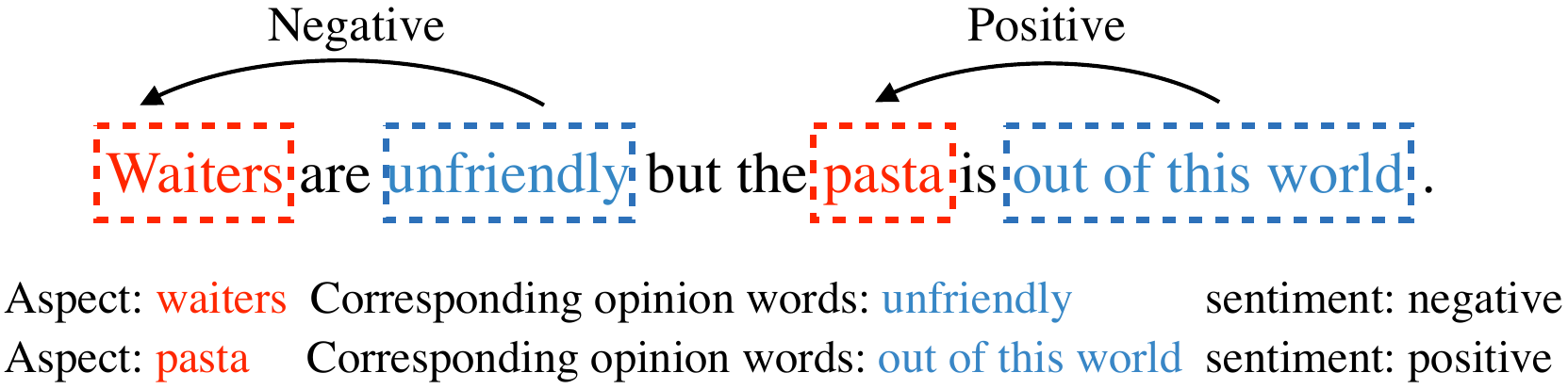}
	\caption{An exmaple for the ALSC task and AOWE task. The words in red are two given aspects. The spans in blue are opinion words. The arrows indicate the correspondence between aspects and opinion words.}
	\label{sentenceexmaple}
\end{figure}

Aspect-level sentiment classification (ALSC) aims to predict sentiment polarity towards a given aspect in a sentence. As Figure~\ref{sentenceexmaple} shows, there are two aspects mentioned in the sentence ``\emph{waiters are unfriendly but the pasta is out of this world.}'', namely ``\emph{waiters}'' and ``\emph{pasta}''. The sentiments expressed towards each aspect are negative and positive respectively. Different from ALSC, aspect-oriented opinion words extraction (AOWE) is a recently proposed ABSA subtask~\cite{fan2019target}. The objective of this task is to extract the corresponding opinion words towards a given aspect from the sentence. Opinion words refer to the word/phrase of a sentence used to express attitudes or opinions explicitly. In the example above, ``\emph{unfriendly}'' is the opinion word towards the aspect ``\emph{waiters}'', and ``\emph{out of this world}'' is the opinion words towards the aspect ``\emph{pasta}''.

It is a common sense that positive opinion words imply positive sentiment polarity, while negative opinion words correspond to negative sentiment polarity. Inspired by this common sense, we can find that the corresponding opinion words toward a given aspect (which AOWE aims at) help infer the corresponding sentiment (which ALSC aims at). Correspondingly, the sentiment determined in ALSC also can provide some clues to help extract polarity-related opinion words for the AOWE task. Therefore, the goals of AOWE and ALSC are mutually indicative and they can benefit each other.

To exploit the above relation of mutual indication, we propose a novel model, Opinion Transmission Network (OTN), to jointly model two tasks of ALSC and AOWE and finally improve them simultaneously. Overall, OTN contains two base modules, namely the attention-based ALSC module and the CNN-based AOWE module, and two tailor-made opinion transmission mechanisms, respectively from AOWE to ALSC and ALSC to AOWE. Specifically, we utilize the extracted results of AOWE as complementary opinions information and inject them into the ALSC module in the form of additional attention. To successfully transmit implicit opinions from ALSC to AOWE, we unearth that the features in attention layer of the ALSC module keep abundant useful aspect-related opinions, which can be utilized to facilitate AOWE. It is worth noting that our proposed model works without requiring simultaneous annotations of AOWE and ALSC on the same data, thus it can be applied in more practical scenarios.

The main contributions of this work can be summarized as follows:
\begin{enumerate}
	\item
	To make full use of high-cost labeled data, we are the first to propose exploiting mutual indication between ALSC and AOWE to improve both tasks.
	\item
	To exploit the connection effectively, we propose a joint neural model Opinion Transmission Network (OTN) with two novel opinion transmission mechanisms. During network training, opinion clues in both modules can flow bi-directionally through the interactions.
	\item
	We conduct experiments and analysis on the benchmark datasets. Experiment results confirm that the performance of ALSC and AOWE can be both improved through our designed opinion transmission mechanisms, and our model outperforms strong baselines on two tasks.
\end{enumerate}

\section{Preliminary}
In this section, we introduce some necessary notations and the task formalizations of the ALSC and AOWE.

\subsection{ALSC Formalization}
ALSC aims to classify the sentiment of a given aspect in a sentence into one set of pre-defined  sentiment categories. Specifically, given a sentence containing $n$ words $s=\{w_1,w_2,...,w_n\}$ and an aspect $w_a$ in $s$ (we notate an aspect as one word $w_a$ for simplicity, and $a$ is the index of the aspect in the sentence), the task is to assign a label $y^{ALSC} \in C$ to an input pair $<s, w_a>$, where $C$ is the set of pre-defined sentiment categories (i.e., positive, negative and neutral).

\subsection{AOWE Formalization}
AOWE aims at extracting the corresponding opinion words towards a given aspect from a sentence. Different from ALSC, it is formalized as an aspect-oriented sequence labeling task~\cite{fan2019target}. Given an input pair $<s, w_a>$, the task is to assign a label $y_i^{AOWE}\in \{B, I, O\}$ for each word $w_i$ in the sentence $s$. The three labels B, I and O refer to the beginning, inside and outside of an aspect, respectively, and they follow the standard BIO notation used in sequence labeling. The spans composed by the tags $B$ and $I$ represent the corresponding opinion words of the aspect $w_a$. It is obvious that a sentence may have different labeling results for different aspects. An example is shown in Table~\ref{aowedefinition}.
\begin{table}[ht]
	\small
	\centering
	\caption{Different labeling results of a sentence when given different aspects. The aspects are highlighted in underline and the opinion words/phrases are in bold.}
	\label{aowedefinition}
	\begin{tabular}{c p{10cm}}
		\hline
		1. & \textrm{\underline{Waiters}/O are/O very/O \textbf{friendly}/B and/O the/O pasta/O is/O out/O of/O this/O world/O ./O} \\
		2. & \textrm{Waiters/O are/O very/O friendly/O and/O the/O \underline{pasta}/O is/O \textbf{out}/B \textbf{of}/I \textbf{this}/I \textbf{world}/I ./O} \\
		\hline
	\end{tabular}
\end{table}

\section{Opinion Transmission Network}
Opinion transmission network (OTN) aims to exploit the connections between ALSC and AOWE to facilitate both tasks. In this section, we first give an overall description of OTN. Then we introduce the base ALSC module and AOWE module in the OTN model. Finally, we present our tailor-made opinion transmission mechanisms to achieve opinions interaction between two modules.

\subsection{Overall Description}
\begin{figure}[ht]
\centering
\includegraphics[width=0.99\textwidth]{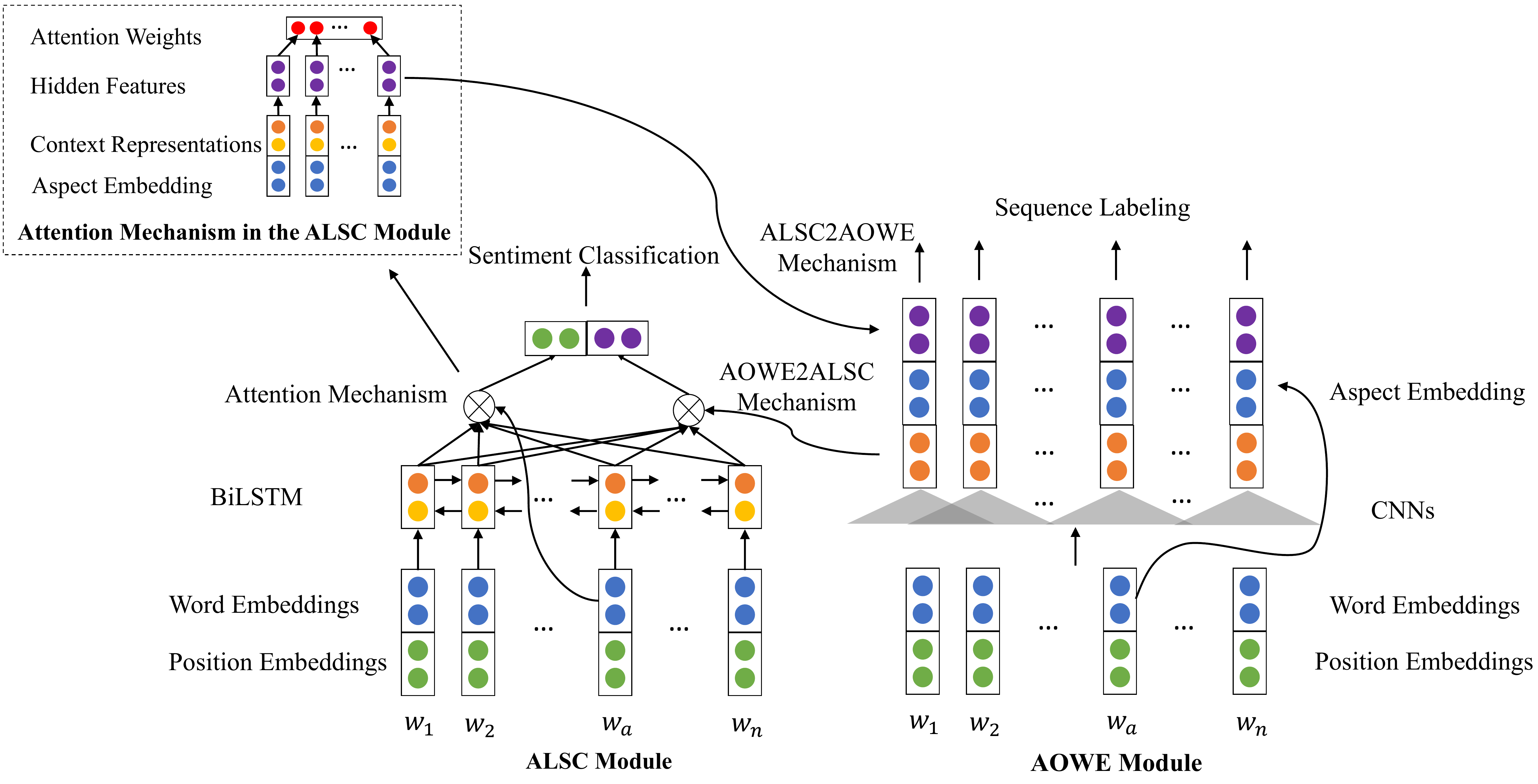}
\caption{Architecture of Opinion Transmission Network.}
\label{fig_model}
\end{figure}

Figure \ref{fig_model} shows the overall architecture of Opinion Transmission Network (OTN). It consists of a base ALSC module and a base AOWE module, as well as bidirectional opinion transmission mechanisms, respectively from AOWE to ALSC (AOWE2ALSC) and ALSC to AOWE (ALSC2AOWE). Following most state-of-the-art works~\cite{wang2016attention,ma2017interactive,gu2018position}, we employ a typical attention-based BiLSTM network as our base ALSC module in this work. In terms of AOWE, we adopt CNN as the base module for two considerations. The first reason is that CNN is widely used in various sequence labeling tasks and achieve state-of-the-art results, such name entity recognition~\cite{zhu2019can}, Chinese words segmentation~\cite{wang-xu-2017-convolutional}, and aspect extraction~\cite{poria2016aspect}. Secondly, it can work in parallel and has more fast computation efficiency. Additionally, we enhance the ALSC module and the AOWE module with position embeddings~\cite{gu2018position} to incorporate aspect information.

For the ALSC task, distinguishing the aspect-related opinion words is helpful to predict the sentiment polarity of the aspect. Thus, we design the opinion transmission mechanism AOWE2ALSC to integrate opinion words information from AOWE into ALSC. Specifically, we transform the prediction results of the AOWE module into the form of auxiliary attention, as additional sentiment evidence for the ALSC module.

The ALSC2AOWE mechanism aims to exploit implicit opinion clues of ALSC to improve the AOWE task. In the ALSC module, the attention weights over words can indicate the aspect-related opinion words, while it is low-dimension and easily ignored when incorporated into the AOWE module. Therefore, we step backward and leverage the intermediate features in the attention layer of the ALSC module as latent opinions to improve the AOWE task.

\subsection{Base ALSC Module}
The base ALSC module is an attention-based BiLSTM network enhanced with position embedding technique. Given a sentence $s=\{w_1, w_2, \cdots, w_n\}$ and an aspect $w_a$ in $s$, we first concatenate the word embedding and position embedding of each word $w_i$ as the word representation $\mathbf{e}_i$, i.e., $\mathbf{e}_i = [\mathbf{E}_{word}(w_i); \mathbf{E}_{pos}(l_i)]$. The $l_i$ indicates the relative distance of the word $w_i$ to the aspect $w_a$ and is calculated as $l_i = |i-a|$. The $\mathbf{E}_{word}$ and $\mathbf{E}_{pos}$ respectively represent the word embedding table and position embedding table. 

With the enhanced word representations $\{\mathbf{e}_1, \mathbf{e}_2, \cdots, \mathbf{e}_n\}$, a BiLSTM network is applied to encode them and generate the corresponding hidden states $\{\mathbf{h}_1, \mathbf{h}_2, \cdots, \mathbf{h}_n\}$. Then we use the aspect representation $\mathbf{e}_a$ as query and employ the attention mechanism to capture potential opinion clues for the ALSC task. The attention weight $\alpha_i$ of the word $w_i$ is defined as:
\begin{align}
\mathbf{h}_i^{a}&=\mathbf{W}_{e}[\mathbf{e}_i;\mathbf{e}_a], \label{attentionfeature}\\
u_i&=\mathbf{v}_{u}\tanh(\mathbf{h}_i^{a}+\mathbf{b}_{u}), \\
\alpha_i &= \frac{\exp (u_i)}{\sum_{j=1}^{n} \exp(u_j) },
\end{align}
where $\mathbf{W}_{e}$ denotes the weight matrix, $\mathbf{v}_{u}$ represents the weight vector, and $\mathbf{b}_{u}$ is the bias.

Finally, the aspect-related sentence representation $\mathbf{r}_{a}$ is a weighted sum of context representations $\mathbf{H} = \{\mathbf{h}_1, \mathbf{h}_2, \cdots, \mathbf{h}_n\}$, i.e., $\mathbf{r}_{a} = \mathbf{H}\boldsymbol{\alpha}$. In the base ALSC module, the representation $\mathbf{r}_{a}$ is fed into a linear layer and a softmax layer to predict the sentiment polarity of the aspect $a$ in the sentence $s$.

\subsection{Base AOWE Module}
Similarily, the word representation $\mathbf{e}_i$ in the base AOWE is obtained by concatenating the word embedding and position embedding of each word $w_i$. We then employ a CNN encoder to capture context information in the sequence $\{\mathbf{e}_1, \mathbf{e}_2, \cdots, \mathbf{e}_n\}$ and obtain the corresponding feature vector $\mathbf{c}_i$ of the word $w_i$:
\begin{equation}
[\mathbf{c}_1, \mathbf{c}_2, \cdots, \mathbf{c}_n]=\mathrm{CNN}([\mathbf{e}_1, \mathbf{e}_2, \cdots, \mathbf{e}_n],\theta_{\text{CNN}}),
\end{equation}
where $\theta_{\text{CNN}}$ represents the parameters of the CNN encoder.

The CNN encoder consists of 5 CNN layers. Each layer has a set of convolution filters, and each filter can map representations of k continuous words to single feature scalar, where k is the kernel size. ReLU activation is applied to each feature vector. We will present the details of the hyperparameters of the base AOWE module in the experiment settings.

To further incorporate the aspect information, we concatenate the CNN feature vector $\mathbf{c}_i$ with the word embedding of the given aspect $a$ as the final representation of each word $w_i$:
\begin{equation}
\mathbf{r}_i^{o}=[\mathbf{c}_i; \mathbf{E}_{word}({w_a})].
\end{equation}

Finally, the sequence representations $\{\mathbf{r}_1^o, \mathbf{r}_2^{o}, \cdots, \mathbf{r}_n^{o}\}$ is fed into a two-layer perceptron and a softmax layer to predict the tag probability distribution for each word in the sentence $s$:
\begin{equation}
\mathbf{\hat y}_i^{AOWE}=\mathrm{softmax}(W_{o1}\mathrm{ReLU}(W_{o2}\mathbf{r}_i^o)+\mathbf{b}_o),
\end{equation}
where $\mathbf{W}_{o1}$ and $\mathbf{W}_{o2}$ are the weight matrices, $\boldsymbol{b}_o$ denotes the bias.

\subsection{Opinion Transmission: AOWE2ALSC}
As we have mentioned, aspect-oriented opinion words in a sentence can provide powerful evidence to infer the corresponding sentiment of the aspect. Therefore, we propose the opinion transmission mechanism AOWE2ALSC to leverage predictions from the AOWE module to help the ALSC module focus on aspect-oriented opinion words, thereby make more comprehensive sentiment predictions.

Specifically, we map the predicted probabilities of BIO tags over each word in the AOWE module to a  probability distribution of each word being aspect-related opinion word as follows:
\begin{equation}
\mathbf{p}=\mathrm{softmax}([\mathbf{\hat y}_1^{AOWE}, \mathbf{\hat y}_2^{AOWE}, \cdots, \mathbf{\hat y}_n^{AOWE}]^T\mathbf{W}_{trans}),
\end{equation}
where $\mathbf{W}_{trans}\in \mathbb{R}^{3\times1}$ is a weight matrix and maps probabilities of each word being tagged with $B,I,O$ to a single score.

Since the probability distribution $\mathbf{p}$ can be regarded as an additional attention knowledge from the AOWE moduel, we also merge the context representations $\mathbf{H}$ in the ALSC module with $\mathbf{p}$:
\begin{equation}
\mathbf{r}_{opinion}=H\mathbf{p}.
\end{equation}

Finally, we concatenate the opinion representation $\mathbf{r}_{opinion}$ with the original representation $\mathbf{r}_a$ in the ALSC module to predict the aspect-level sentiment:
\begin{equation}
\mathbf{\hat y}^{ALSC}=\mathrm{softmax}(\mathbf{W}_a[\mathbf{r}_{a}; \mathbf{r}_{opinion}]+\mathbf{b}_a)
\end{equation}
where $\mathbf{W}_a$ is the weight matrix and $\mathbf{b}_a$ denotes the bias.

\subsection{Opinion Transmission: ALSC2AOWE}
The base ALSC module can capture some latent aspect-related opinion words through the attention mechanism. However, the attention weight $\alpha_i$ is the 1-dimension scalar in the ALSC module and easily neglected when we use it to enhance the AOWE module. Therefore, we exploit the attention feature $\mathbf{h}_i^{a}$ in Equation~\ref{attentionfeature} to enrich the context representations of the ALSC module as follows:
\begin{equation}
\mathbf{r}_i^{o'}=[\mathbf{r}_i^{o}; \mathbf{h}_i^{a}].
\end{equation}

Finally, the enriched context representation $\mathbf{r}_i^{o'}$ is used to the predict tag of the word $w_i$ for the AOWE task:
\begin{equation}
\mathbf{\hat y}_i^{AOWE}=\mathrm{softmax}(W_{o1}\mathrm{ReLU}(W_{o2}\mathbf{r}_i^{o'})+\mathbf{b}_o),
\end{equation}

\subsection{Training}
For the ALSC task, we use cross-entropy loss between predicted sentiment label and the gold sentiment label as the task loss, which is defined as follows:
\begin{equation}
L^{ALSC}=- \sum_{d\in D} \sum_{i=1}^{|C|}\mathbb{I}(y^{ALSC}=i) \log \hat y_i^{ALSC},
\end{equation}
where $D$ indicates all data sample, $C$ denotes the sentiment label set, and $\hat y_i^{ALSC}$ is the predicted probability of the input sample belonging to the $i$-th sentiment.

In terms of the AOWE task, we define the cross-entropy loss as follows:
\begin{equation}
L^{AOWE}=- \sum_{d\in D} \sum_{i=1}^n\sum_{j=0}^2 \mathbb{I}(y_i^{AOWE}=j)\hat y_{i,j}^{AOWE},
\end{equation}
here the tags $\{O, B, I\}$ are correspondingly converted into labels $\{0,1,2\}$, and $\hat y_{i,j}^{AOWE}$ denotes the probability that the $i$-th word is predicted as the label $j$.

Because OTN is a joint model for both tasks of ALSC and AOWE, we minimize the losses $L^{ALSC}$ and $L^{AOWE}$ iteratively to optimize the OTN model.

\section{Experiments}
\subsection{Datasets and Metrics}
As aforementioned, the OTN model is a joint model without requiring strict annotations on the same data for the ALSC task and the AOWE task. To verify this, we respectively use the datasets 14res for ALSC and 16res for AOWE. They are respectively derived from SemEval Challenge 2014 task 4 \cite{pontiki-etal-2014-semeval} and SemEval Challenge 2016 task 5 \cite{pontiki2016semeval}. The original SemEval datasets do not provide the annotations of the corresponding opinion words for each aspect. Therefore,~\cite{fan2019target} annotate aspect-related opinion words for each sample and remove the samples without containing opinion words. Table~\ref{datasets} shows the statistics of the two datasets, the ``Opinion'' and ``Pair'' respectively denote the number of opinion words and pairs of aspects and opinion words in Table~\ref{datasets}.

\begin{table}[htbp]
    \centering
     \caption{Statistics of ALSC dataset 14res and AOWE dataset 16res.}
    \label{datasets}
\begin{tabular}{cc|cccc||cc|cccc}
\hline
\multicolumn{2}{c|}{ALSC}  & Pos. & Neu. & Neg. & Total & \multicolumn{2}{c|}{AOWE}  & Sentence & Aspect & Opinion & Pair \\ \hline
\multirow{2}{*}{14res} & Train & 2,164  & 633    & 805    & 3,602 & \multirow{2}{*}{16res} & Train & 1,079       & 1,512   & 1,661    & 1,770 \\
                       & Test  & 728    & 196    & 196    & 1,120 & & Test  & 329         & 457  & 485      & 525     \\ \hline
\end{tabular}
\end{table}

We adopt widely-used evaluation metrics for the two tasks. For ALSC, we use accuracy and macro-F1 score as evaluation metrics~\cite{chen2017recurrent,he-etal-2018-exploiting}. For AOWE, we follow the previous work~\cite{fan2019target} and use precision, recall, and F1-score to measure the performance of different methods. An opinion word/phrase is deemed to be correct on the condition that the starting and ending positions of the prediction are both the same as those of the golden word/phrase.

\subsection{Experiment Settings}
We use 300-dimension GloVe~\cite{pennington2014glove} word embeddings pre-trained from 840B tokens to initialize word vectors, which are fixed during the training stage. The position embeddings are 100-dimension vectors and randomly initialized by a uniform distribution $U(-0.01, 0.01)$. The dimension of the LSTM cells is 400. Table~\ref{table_cnn_param} shows the hyperparameters of the CNNs in the AOWE module. We adopt dropout on the embedding layer and the output layer with probability 0.5. Adam optimizer~\cite{kingma2014adam} is applied to update model parameters. The initial learning rate is 1e-3 and the mini-batch size is 16. We randomly select 20\% samples from training sets as the validation sets for tuning hyperparameters and early stopping. We report the average results of 5 repeated experiments for each model.

\begin{table}[htbp]
\centering
\caption{Hyperparameters of the CNNs in the AOWE module.}
\label{table_cnn_param}
\begin{tabular}{c|c|c}
\hline
Layer No.          & Filter length $k$ & Filter numbers \\ \hline
1                  & 1             & 600       \\ \hline
\multirow{3}{*}{2} & 2             & 200       \\
                   & 3             & 200       \\
                   & 4             & 200       \\ \hline
3                  & 5             & 600       \\ \hline
4                  & 5             & 600       \\ \hline
5                  & 5             & 600       \\ \hline
\end{tabular}
\end{table}

\subsection{Compared Methods}

We compare our OTN model with the following methods for ALSC and AOWE.

\subsubsection{ALSC}
We divide the compared ALSC methods into three groups for brevity.

\begin{itemize}
	\item ATAE-LSTM~\cite{wang2016attention}, IAN~\cite{ma2017interactive} and PBAN~\cite{gu2018position} are attention-based methods.
	
	\item MemNN~\cite{tang-etal-2016-aspect}, RAM~\cite{chen2017recurrent}, and CEA~\cite{yang2018multi}, DAuM~\cite{zhu2018enhanced} are all memory-based metheds.
	\item GCAE is a CNN-based model. It proposes a novel Gated Tanh-ReLU Units to selectively output the sentiment features according to the given aspect~\cite{xue2018aspect}.
\end{itemize}

Besides, we also report the performance of our base ALSC module. It is an attention-based BiLSTM network enhanced with position embedding.

\subsubsection{AOWE}We compare our method with five baselines for AOWE.

\begin{itemize}
	\item Dependency-rule uses POS tags of the dependency path between aspect and opinion words from training set as rule templates to detect the corresponding opinion words for the given aspects \cite{zhuang2006movie}.
	
	\item BiLSTM uses word embedding to represent words, then employs a BiLSTM network to capture context information. Finally, the context representations are used to predict the tags for words.
	
	\item PE-BiLSTM employs additional position embeddings based on BiLSTM to represent relative positions of words to the given aspect~\cite{DBLP:conf/aaai/WuZDHC20}.
	
	\item IOG employs six different positional and directional LSTM networks to extract aspect-related opinion words and achieves state-of-the-art resutls~\cite{fan2019target}.

\end{itemize}

\subsection{Main Results}
\begin{table}[!htbp]
	\centering
	\caption{Main experiment results (\%). Best results are in bold.}
	\label{table_main_results}
	\begin{tabular}{c|cc|ccc}
		\hline
		\multirow{2}{*}{Model} & \multicolumn{2}{c|}{ALSC}       & \multicolumn{3}{c}{AOWE}                        \\ \cline{2-6} 
		& Acc.           & F1             & P              & R              & F1             \\ \hline
		ATAE-LSTM              & 78.38          & 66.36          & -              & -              & -              \\
		IAN                    & 78.71          & 67.71          & -              & -              & -              \\
		PBAN                   & 78.62          & 67.45          & -              & -              & -              \\
		MemNN                  & 77.69          & 67.53          & -              & -              & -              \\
		RAM                    & 78.41          & 68.52          & -              & -              & -              \\
		CEA                    & 78.44          & 66.78          & -              & -              & -              \\
		DAuM                   & 77.91          & 66.47          & -              & -              & -              \\
		GCAE                   & 76.09          & 63.29          & -              & -              & -              \\ \hline
		Dependency-rule        & -              & -              & 76.03          & 56.19          & 64.62          \\
		BiLSTM        & -              & -              & 68.68          & 70.51          & 69.57          \\
		PE-BiLSTM              & -              & -              & 82.27          & 74.95          & 78.43          \\ 
		IOG                    & -              & -              & 84.36          & 79.08          & 81.60          \\
		IOG+CRF                    & -              & -              & 84.41          & 79.43          & 81.84          \\ \hline
		Base ALSC module       & 78.85          & 67.69          & -              & -              & -              \\
		Base AOWE module       & -              & -              & 82.83          & \textbf{83.25} & 83.03          \\
		OTN                    & \textbf{79.50} & \textbf{69.08} & \textbf{86.78} & 81.11          & \textbf{83.83} \\ \hline
	\end{tabular}
\end{table}
The main experiment results of ALSC and AOWE are shown in Table \ref{table_main_results}.

In terms of the ALSC task, the performance of attention-based and memory-based methods are comparable, while the CNN-based method GCAE performs worst. This shows the importance of modeling long-term dependency for ALSC. Thus we adopt an attention-based BiLSTM as our base ALSC module. By employing position embedding to incorporate aspect information, the base module achieves very competitive performance. Compared to the base module, our joint model OTN achieves further improvements through leveraging the additional opinion information from the AOWE task.

As for the  AOWE task, the methods Dependency-rule and BiLSTM both perform poorly. The former uses coarse-grain POS patterns and lacks robustness. The latter fails to consider aspect information and outputs the same results for different aspects in a sentence.  In contrast, the aspect-dependent methods PE-BiLSTM and IOG obtain obvious improvements. With the help of CRF, IOG+CRF achieves minor improvements against IOG. Different from LSTM-based methods, our designed base module using CNN and incorporating aspect information produces very competitive results on the AOWE dataset. Nevertheless, our joint model OTN still outperforms it by 0.8\% in F1-score.

Benefiting from the two tailor-made opinion transmission mechanisms, OTN performs better than the two base modules, which proves the existence of mutual indication between the ALSC and AOWE tasks. Besides, OTN consistently outperforms other compared methods in both tasks. The comparison validates the effectiveness of our model.

\subsection{Ablation Study}

\begin{table}[!htbp]
	\centering
	\caption{The experiment results of ablation study.}
	\label{table_ablation}
	\begin{tabular}{c|cc|ccc}
		\hline
		\multirow{2}{*}{Model} & \multicolumn{2}{c|}{ALSC} & \multicolumn{3}{c}{AOWE} \\ \cline{2-6} 
		& Acc.        & F1          & P       & R      & F1     \\ \hline
		OTN                & 79.50       & 69.08       & 86.78   & 81.11  & 83.83  \\ \hline
		-ALSC task                & -           & -           & 86.56   & 80.98  & 83.64  \\
		-AOWE task                & 78.62       & 67.36       & -       & -      & -      \\
		-AOWE2ALSC         & 79.31       & 68.89       & 85.86   & 81.59  & 83.65  \\
		-ALSC2AOWE         & 78.94       & 67.74       & 81.67   & 82.47  & 82.05  \\ \hline
	\end{tabular}
\end{table}
To investigate the effects of the two opinion transmission mechanisms on OTN, we also conduct ablation study. In the experiments of ``-ALSC task'' and ``-AOWE task'', we keep the model architecture unchanged but respectively remove the ALSC data for ``-ALSC task'' and AOWE data for ``-AOWE task'. The ``-AOWE2ALSC'' and ``-ALSC2AOWE'' indicate that we remove the AOWE2ALSC mechanism or ALSC2AOWE mechanism from OTN.

Table \ref{table_ablation} shows the results of ablation study. We can observe the performance of the model drops when it is trained on a single task or without opinion transmission mechanisms. The observation proves that OTN exploits the connection between the ALSC and AOWE tasks successfully and achieves improvements through our proposed opinion transmission mechanisms.

\section{Related Work}
\subsection{ALSC}
Most recent ALSC research utilizes the attention-based networks to capture the latent sentiment clues from the sentence for the given aspect, such as ATAE-LSTM~\cite{wang2016attention}, IAN~\cite{ma2017interactive} and PBAN~\cite{gu2018position}, etc. On this basis, \cite{tang-etal-2016-aspect} employs memory network to conduct multi-hop attention to obtain more powerful sentiment clues for detecting the sentiment polarity of the aspect. Following the idea, memory-based methods achieve competitive performance on the ALSC~\cite{tang-etal-2016-aspect,chen2017recurrent,yang2018multi,zhu2018enhanced}. In addition, CNN~\cite{xue2018aspect}, capsule network~\cite{chen2019transfer}, and additional document sentiment data~\cite{he-etal-2018-exploiting} are also applied for this task.

\subsection{AOWE}
AOWE is a relatively new ABSA subtask.~\cite{fan2019target} formalizes it as an aspect-oriented sequence labeling task, and designs a state-of-the-art sequence labeling model based on LSTMs. Before them, a few works focus on the pair of aspect and opinion words.~\cite{hu2004mining} proposes a rule-mining method to extract aspect words and regards the nearest adjective of aspect as the corresponding opinion words.~\cite{zhuang2006movie} uses dependency-tree templates to extract valid aspect-opinion pairs.

\section{Conclusion}
In ABSA research, Aspect-level sentiment classification (ALSC) and aspect-oriented opinion words extraction (AOWE) are two highly relevant tasks. Previous works usually focus on one of the two tasks and neglect mutual indication between them. In this paper, we propose a novel joint model, Opinion Transmission Network (OTN), to exploit the potential connection between ALSC and AOWE to benefit them simultaneously. In OTN, two tailor-made opinion transmission mechanisms are designed to control opinion clues to flow respectively from ALSC to AOWE and AOWE to ALSC. Experiment results on two tasks validate the effectiveness of our method.

\subsubsection{Acknowledgements.}
This work was supported by the NSFC (No. 61976114, 61936012) and National Key R\&D Program of China (No. 2018YFB1005102).

%
%
%
\bibliographystyle{splncs04}
\bibliography{nlpcc}

\begin{thebibliography}{10}
\providecommand{\url}[1]{\texttt{#1}}
\providecommand{\urlprefix}{URL }
\providecommand{\doi}[1]{https://doi.org/#1}

\bibitem{chen2017recurrent}
Chen, P., Sun, Z., Bing, L., Yang, W.: Recurrent attention network on memory
  for aspect sentiment analysis. In: EMNLP. pp. 452--461 (2017)

\bibitem{chen2019transfer}
Chen, Z., Qian, T.: Transfer capsule network for aspect level sentiment
  classification. In: ACL. pp. 547--556 (2019)

\bibitem{fan2019target}
Fan, Z., Wu, Z., Dai, X., Huang, S., Chen, J.: Target-oriented opinion words
  extraction with target-fused neural sequence labeling. In: NAACL. pp.
  2509--2518 (2019)

\bibitem{gu2018position}
Gu, S., Zhang, L., Hou, Y., Song, Y.: A position-aware bidirectional attention
  network for aspect-level sentiment analysis. In: COLING. pp. 774--784 (2018)

\bibitem{he-etal-2018-exploiting}
He, R., Lee, W.S., Ng, H.T., Dahlmeier, D.: Exploiting document knowledge for
  aspect-level sentiment classification. In: ACL. pp. 579--585 (Jul 2018)

\bibitem{hu2004mining}
Hu, M., Liu, B.: Mining and summarizing customer reviews. In: ACM SIGKDD. pp.
  168--177 (2004)

\bibitem{kingma2014adam}
Kingma, D.P., Ba, J.: Adam: {A} method for stochastic optimization. In: ICLR
  (2015)

\bibitem{ma2017interactive}
Ma, D., Li, S., Zhang, X., Wang, H.: Interactive attention networks for
  aspect-level sentiment classification. In: IJCAI. pp. 4068--4074 (2017)

\bibitem{pennington2014glove}
Pennington, J., Socher, R., Manning, C.D.: Glove: Global vectors for word
  representation. In: EMNLP. pp. 1532--1543 (2014)

\bibitem{pontiki2016semeval}
Pontiki, M., Galanis, D., Papageorgiou, H., Androutsopoulos, I., Manandhar, S.,
  Al-Smadi, M., Al-Ayyoub, M., Zhao, Y., Qin, B., De~Clercq, O., et~al.:
  Semeval-2016 task 5: Aspect based sentiment analysis. In: (SemEval 2016)
  (2016)

\bibitem{pontiki-etal-2014-semeval}
Pontiki, M., Galanis, D., Pavlopoulos, J., Papageorgiou, H., Androutsopoulos,
  I., Manandhar, S.: {S}em{E}val-2014 task 4: Aspect based sentiment analysis.
  In: ({S}em{E}val 2014). pp. 27--35 (Aug 2014)

\bibitem{poria2016aspect}
Poria, S., Cambria, E., Gelbukh, A.: Aspect extraction for opinion mining with
  a deep convolutional neural network. Knowledge-Based Systems  \textbf{108},
  42--49 (2016)

\bibitem{tang-etal-2016-aspect}
Tang, D., Qin, B., Liu, T.: Aspect level sentiment classification with deep
  memory network. In: EMNLP. pp. 214--224. Association for Computational
  Linguistics, Austin, Texas (Nov 2016)

\bibitem{wang-xu-2017-convolutional}
Wang, C., Xu, B.: Convolutional neural network with word embeddings for
  {C}hinese word segmentation. In: IJCNLP. pp. 163--172 (Nov 2017)

\bibitem{wang2016attention}
Wang, Y., Huang, M., Zhu, X., Zhao, L.: Attention-based lstm for aspect-level
  sentiment classification. In: EMNLP. pp. 606--615 (2016)

\bibitem{DBLP:conf/aaai/WuZDHC20}
Wu, Z., Zhao, F., Dai, X., Huang, S., Chen, J.: Latent opinions transfer
  network for target-oriented opinion words extraction. In: AAAI. pp.
  9298--9305 (2020)

\bibitem{xue2018aspect}
Xue, W., Li, T.: Aspect based sentiment analysis with gated convolutional
  networks. In: ACL. pp. 2514--2523 (2018)

\bibitem{yang2018multi}
Yang, J., Yang, R., Wang, C., Xie, J.: Multi-entity aspect-based sentiment
  analysis with context, entity and aspect memory. In: AAAI (2018)

\bibitem{zhu2018enhanced}
Zhu, P., Qian, T.: Enhanced aspect level sentiment classification with
  auxiliary memory. In: COLING. pp. 1077--1087 (2018)

\bibitem{zhu2019can}
Zhu, Y., Wang, G.: Can-ner: Convolutional attention network for chinese named
  entity recognition. In: NAACL. pp. 3384--3393 (2019)

\bibitem{zhuang2006movie}
Zhuang, L., Jing, F., Zhu, X.Y.: Movie review mining and summarization. In:
  CIKM. pp. 43--50 (2006)

\end{thebibliography}

\end{document}